\documentclass[conference]{IEEEtran}
\IEEEoverridecommandlockouts
\usepackage{cite}
\usepackage{amsmath,amssymb,amsfonts}
\usepackage{algorithmic}
\usepackage{graphicx}
\usepackage{textcomp}
\usepackage{xcolor}
\usepackage{pifont}
\def\BibTeX{{\rm B\kern-.05em{\sc i\kern-.025em b}\kern-.08em
    T\kern-.1667em\lower.7ex\hbox{E}\kern-.125emX}}
\definecolor{cvprblue}{rgb}{0.21,0.49,0.74}
\usepackage{multirow}
\usepackage{graphicx}
\usepackage{verbatim}
\usepackage{arydshln}
\usepackage{array}
\usepackage{xcolor}
\usepackage{multirow}
\usepackage{colortbl}
\usepackage{booktabs}
\usepackage[hidelinks]{hyperref}
\definecolor{gaincolor}{RGB}{34, 139, 34} 
\definecolor{dropcolor}{RGB}{218, 61, 61} 
\definecolor{ourscolor}{RGB}{235, 250, 235} 
\definecolor{headergray}{RGB}{240, 240, 240} 
\newcommand{\cmark}{\ding{51}} 
\newcommand{\xmark}{\ding{55}} 

\definecolor{mygray}{gray}{.9}

\newcommand{\pub}[1]{~{\color{gray}\fontsize{6pt}{7pt}\selectfont [#1]}}

\begin{document}

\title{Knowledge-Preserved Model Tuning in Null-Space for Robust Spatio-Temporal Video Grounding}
\author{
Haoxuan Chen\textsuperscript{1}, Xianqin Liu\textsuperscript{2} and Jian-Fang Hu\textsuperscript{1,3,4\dag} \\
\textsuperscript{1}School of Computer Science and Engineering, Sun Yat-sen University, China\\
\textsuperscript{2}National Information Center of GACC (Guangdong), GuangZhou, China\\
\textsuperscript{3}Guangdong Province Key Laboratory of Information Security Technology, China\\
\textsuperscript{4}Key Laboratory of Machine Intelligence and Advanced Computing, Ministry of Education, China\\
chenhx253@mail2.sysu.edu.cn, liuxianqin007@hotmail.com, hujf5@mail.sysu.edu.cn
}
\maketitle
\renewcommand{\thefootnote}{}
\footnotetext{\dag\ Corresponding Author}
\maketitle
\begin{abstract}
Spatio-Temporal Video Grounding aims to localize object tubes based on textual queries. While recent methods have achieved remarkable success, they mainly focus on high-quality(HQ) inputs, neglecting the widespread presence of low-quality(LQ) videos in real-world scenarios. Although tuning methods like LoRA can adapt to degraded inputs, they inevitably disrupt pre-trained knowledge. To address this, we propose Null-Space Tuning (NST). This framework exploits the geometric property that adding vectors within the null-space of frozen weights to the layer input does not affect the output. Leveraging this, NST injects learnable residuals into input features that can be selectively invisible to the pre-trained backbone. Specifically, NST combines the Quality-Adaptive Unit and Dual-Space Reparameterization to synthesize these residuals by confining components for HQ inputs to the null-space, while directing restoration components for LQ inputs to the non-null space. As the frozen weights eliminate null-space components, we effectively rectify degraded inputs while preserving pre-trained knowledge for HQ inputs. Extensive experiments show that NST outperforms state-of-the-art methods on our Mixed-Quality benchmark.

\end{abstract}


\section{Introduction}
\label{sec:intro}
Spatio-Temporal Video Grounding (STVG) aims to localize target object tubes in untrimmed videos based on natural language queries~\cite{where}. Driven by the advancements in DETR-style architectures~\cite{carion2020end}, recent methods~\cite{gu2024context, gu2025knowing} have achieved impressive performance by modeling fine-grained cross-modal interactions. However, these approaches mainly assume that input videos are of high-quality. In real-world scenarios, videos inevitably suffer from degradations such as motion blur, defocus blur, or compression artifacts. These degradations lead to a severe feature distribution shift, where the local visual details essential for grounding, such as object boundaries, become inconsistent with the patterns learned by pre-trained models. Consequently, standard models struggle to align degraded visual features with textual queries, leading to significant localization failures on low-quality inputs.


To bridge this gap, a straightforward solution is to apply Parameter-Efficient Fine-Tuning (PEFT) methods like LoRA~\cite{hu2022lora}. However, standard PEFT employs a static injection strategy where learned parameters are fixed and applied uniformly to all inputs regardless of their quality. This causes a dilemma in real-world scenarios: since both LQ and HQ features share the same projection space, adjusting parameters to restore degraded inputs inevitably disrupts the well-learned representations of HQ data, resulting in catastrophic forgetting.

To resolve this dilemma, we propose Null-Space Tuning. This framework leverages the geometric redundancy of pre-trained weights to find a latent subspace where adaptation signals within it remain invisible to the pre-trained backbone. Mathematically, this corresponds to the subspace orthogonal to the active weights, formally defined as the \textit{Null-Space}. By restricting updates for HQ inputs to this null-space, we effectively prevent these updates from altering the layer outputs. These updates are injected into the input features as residuals, enabling knowledge-preserved tuning. Specifically, we design a Quality-Adaptive Unit (QAU) that leverages a multi-modal reference bank to assess the semantic gap in degraded frames and generate the corresponding restoration coefficients. Furthermore, we introduce Dual-Space Reparameterization to synthesize the final residual as a linear combination of orthogonal bases derived from the SVD of the frozen weights. Driven by the coefficients from QAU, this mechanism constructs the residual using null-space bases for HQ inputs to preserve original outputs, while switching to non-null space bases for LQ inputs to enable restoration. The residual is then added into the input features. As the backbone eliminates residuals within null-space, this selective tuning method ensures effective restoration for LQ inputs while preserving HQ knowledge during the forward pass. Experiments on our Mixed-Quality benchmark show that our method effectively restores degraded inputs without compromising pre-trained knowledge.

Our contributions can be summarized as follows:
\begin{itemize}
    \item We propose Null-Space Tuning, a novel geometric tuning framework that resolves the conflict between adapting to low-quality inputs and preserving pre-trained knowledge on high-quality data.
    \item We design a Quality-Adaptive Unit that utilizes a multi-modal reference bank to generate semantic-aware residuals, combined with a Dual-Space Reparameterization to selectively inject residuals based on input qualities.
    \item We construct a Mixed-Quality STVG Benchmark simulating real-world video degradations. Extensive experiments on it demonstrate that NST achieves superior robustness, outperforming state-of-the-art PEFT methods in restoration gain and knowledge preservation.
\end{itemize}

\section{Related Work}
\label{sec:related_work}
 \begin{figure*}[t!]
    \centering 
    \includegraphics[width=\linewidth]{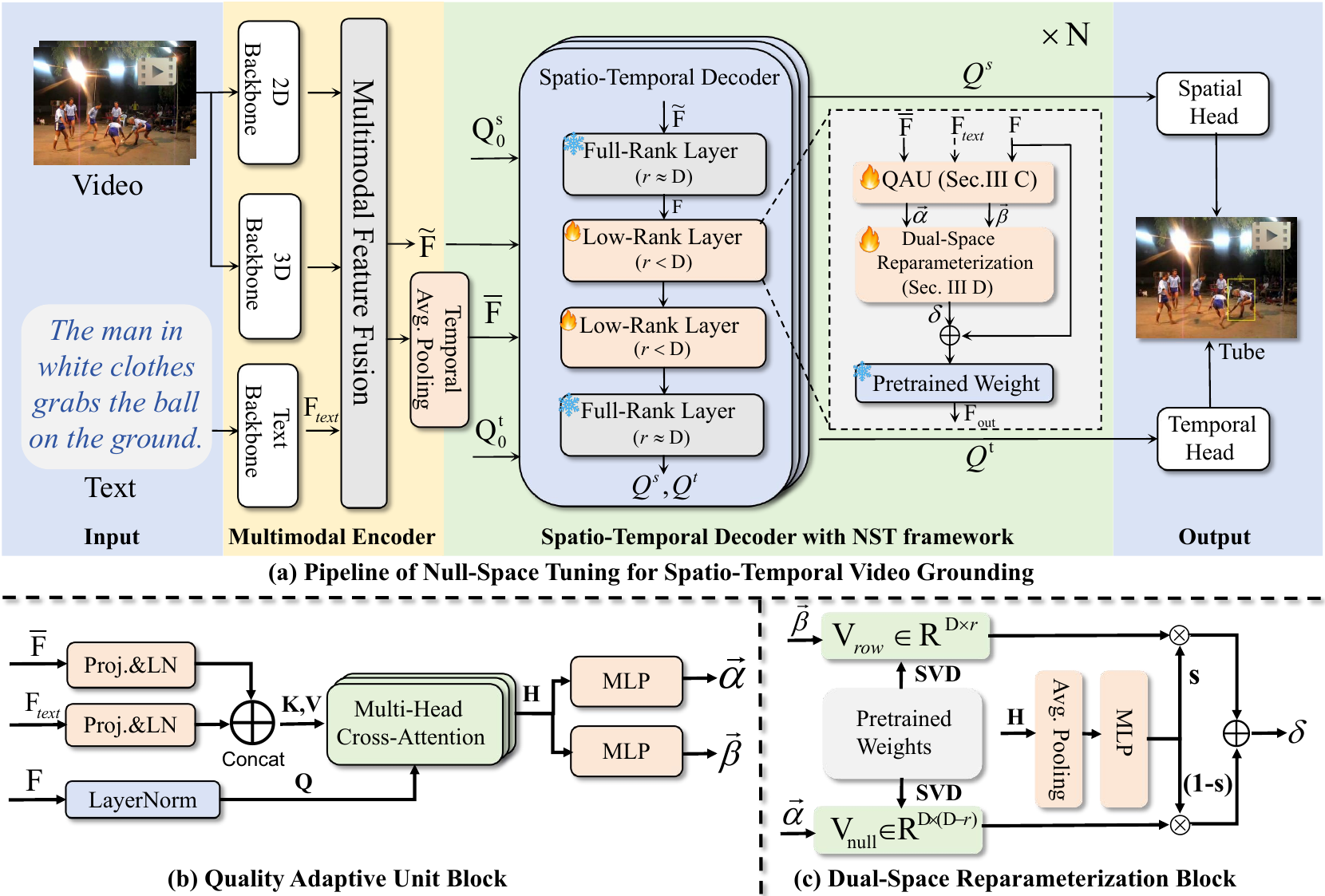}
    \vspace{-7mm}
    \caption{\textbf{Overview of Null-Space Tuning.} 
    (a) Pipeline: We freeze the backbone and insert our modules into the decoder layers with low rank.
    (b) Quality-Adaptive Unit: Constructs a reference bank from global visual $\tilde{F}$ and textual features $F_{text}$, utilizing degraded local frames $F$ as queries to retrieve missing semantics via Cross-Attention.
    (c) Dual-Space Reparameterization: Constructs residuals via orthogonal bases. A router $s$ directs restoration signals to the row-space ($\mathbf{V}_{row}$) for LQ inputs, while constraining noise within the null-space ($\mathbf{V}_{null}$) for HQ inputs to minimize interference.}
    \label{fig:framework}
    \vspace{-5mm}
\end{figure*} 
\subsection{Spatio-Temporal Video Grounding}
STVG aims to localize object tubes based on textual queries. Early methods~\cite{where, zhang2020object, tang2021human} adopted a two-stage paradigm with proposal generation and ranking, which suffers from error propagation. The field has since evolved to one-stage Transformer architectures, where methods like STVGBert~\cite{su2021stvgbert} and TubeDETR~\cite{yang2022tubedetr} leverage end-to-end tube generation to establish a direct localization paradigm. Recent SOTA models~\cite{embrace, gu2024context, gu2025knowing} have further pushed performance boundaries by modeling fine-grained cross-modal interactions. However, these methods mainly work under the assumption of high-quality inputs, lacking mechanisms to handle real-world video degradations. Unlike prior methods, we propose a tuning framework that explicitly recovers these missing semantics to enable robust localization on mixed-quality inputs.  


\subsection{Parameter-Efficient Fine-Tuning}
PEFT adapts models by optimizing minimal parameters while freezing the backbone. Existing methods generally fall into additive approaches (e.g., Adapters~\cite{houlsby2019parameter}, Prompt Tuning~\cite{jia2022visual}) or reparameterization strategies (e.g., LoRA~\cite{hu2022lora}). While recent variants like AdaLoRA~\cite{zhang2023adalora} and DoRA~\cite{liu2024dora} further optimize rank allocation or weight decomposition, they fundamentally rely on a static injection mechanism, where learned parameters remain fixed for all samples during inference. This input-agnostic nature limits their capability when handling mixed-quality data in real world scenarios. In contrast, our method introduces an instance-aware strategy, leveraging the null-space to dynamically route features based on input quality without altering the frozen model.

\subsection{Null-Space in Deep Learning}
Null-space analysis is pivotal in linear algebra and has been adopted in continual learning~\cite{wang2021training, fang2024alphaedit, cheng2024mamba} to prevent forgetting by constraining gradient updates orthogonal to previous tasks. Unlike previous works that use null space solely for gradient projection or static regularization, we exploit the null-space of frozen weights to dynamically tune model. By decomposing the feature space via SVD, we establish a dynamic tuning method where restoration updates for low-quality inputs are directed to the non-null space, whereas potential noise from high-quality inputs is mathematically constrained within the null-space to preserve pre-trained knowledge.

\section{Method}
\label{sec:method}
\subsection{Overview}
We propose Null-Space Tuning, a robust tuning framework that leverages the null-space structure of frozen weights to perform selective tuning. As illustrated in Fig.~\ref{fig:framework}(a), NST integrates two core modules into the decoder layers with lower ranks: 
(1) A Quality-Adaptive Unit (Sec.~\ref{sec:qau}), which estimates restoration coefficients by detecting semantic gaps; 
and (2) A Dual-Space Reparameterization mechanism (Sec.~\ref{sec:reparam}), which constructs residuals using orthogonal bases derived from the SVD of frozen weights. By routing restoration residuals for LQ inputs to the row-space while projecting noise into the null-space for HQ inputs, NST achieves selective adaptation while preserving pre-trained knowledge.
\subsection{Problem Formulation}
Let $\mathbf{F} \in \mathbb{R}^{N \times D}$ denote the input features consisting of $N$ tokens to a pre-trained layer with frozen weights $\mathbf{W} \in \mathbb{R}^{D \times D}$. In standard settings, $\mathbf{W}$ is optimized for HQ inputs $\mathbf{F}_{\text{HQ}}$. However, real-world degradations cause a distribution shift, yielding LQ features $\mathbf{F}_{\text{LQ}}$. Directly updating $\mathbf{W}$ to fit $\mathbf{F}_{\text{LQ}}$ alters the projection of $\mathbf{F}_{\text{HQ}}$, causing catastrophic forgetting. To resolve this, we freeze $\mathbf{W}$ and inject a learnable residual $\delta = \mathcal{G}(\mathbf{F})$ into $F$ to construct rectified features $\hat{\mathbf{F}} = \mathbf{F} + \delta$. 
To ensure $\delta$ restores LQ features without disturbing HQ ones, we leverage the geometric redundancy of the frozen weight $\mathbf{W}$. We perform SVD on $\mathbf{W} = \mathbf{U}\mathbf{\Sigma}\mathbf{V}^\top$ to obtain the orthogonal decomposition of the input domain $\mathbb{R}^D$. The right singular vectors $\mathbf{V}$ are partitioned into two orthogonal subspaces: (1) \textit{Row-Space} ($\mathbf{V}_{row}$, also referred to as the non-null space), which corresponds to non-zero singular values and contains directions that modulate the layer output; and (2) \textit{Null-Space} ($\mathbf{V}_{null}$), which corresponds to zero singular values. By definition, any vector $\mathbf{v}$ within this null-space is nullified by the weights. Based on this, we construct $\delta$ to act as a geometric gate. For HQ inputs, we constrain $\delta$ within $\mathbf{V}_{null}$. Since the frozen weights map null-space vectors to zero, the layer output remains unchanged:
\begin{equation}
    \mathbf{W}(\mathbf{F} + \delta) = \mathbf{W}\mathbf{F} + \underbrace{\mathbf{W}\delta}_{=\mathbf{0}} = \mathbf{W}\mathbf{F},
\end{equation}
thereby strictly preserving the pre-trained capabilities. Conversely, for LQ inputs, we route restoration residuals into $\mathbf{V}_{row}$, yielding $\mathbf{W}(\mathbf{F} + \delta) = \mathbf{W}\mathbf{F} + \mathbf{W}\delta$, where the non-zero term $\mathbf{W}\delta$ explicitly modifies the degraded features.

\subsection{Quality-Adaptive Unit}
\label{sec:qau}
To guide the construction of $\delta$, we introduce the QAU to detect semantic gaps and generate candidate restoration signals. As shown in Fig.~\ref{fig:framework}(b), QAU acts as a dual-path generator to prepare restorative residuals.

Specifically, input features $\mathbf{F}$ derived from degraded frames often contain noise that corrupts local details. To rectify this, we construct a robust reference bank $\mathbf{M}$ as:
\begin{equation}
    \mathbf{M} = [\text{LN}(\text{Proj}(\bar{\mathbf{F}})); \ \text{LN}(\text{Proj}(\mathbf{F}_{text}))],
\end{equation}
where $\bar{\mathbf{F}}$ is the global visual context obtained via temporal pooling to capture a stable visual representation, and $\mathbf{F}_{text}$ is the textual features providing clean semantic guidance. Subsequently, we employ Multi-Head Cross-Attention(MHCA) to query this bank, treating degraded input frames $\mathbf{F}$ as Queries to retrieve missing semantics:
\begin{equation}
    \mathbf{H} = \text{MHCA}(\mathbf{Q}=\text{LN}(\mathbf{F}), \mathbf{K}=\mathbf{M}, \mathbf{V}=\mathbf{M} ),
\end{equation}
thereby allowing the model to selectively retrieve missing semantics from the robust reference to compensate for local degradation. Finally, the retrieved features $\mathbf{H}$ are mapped into dual projection coefficients of null space and row space via two parallel MLPs:
\begin{equation}
    \vec{\alpha} = \text{MLP}_{null}(\mathbf{H})\qquad
    \vec{\beta}  = \text{MLP}_{row}(\mathbf{H}) ,
\end{equation}
where $\vec{\beta} \in \mathbb{R}^{N \times r}$ carries the restoration information intended to be injected via row-space and $\vec{\alpha} \in \mathbb{R}^{N \times (D-r)}$ collects the noise intended to be discarded via null-space. By separating the information into these two components, we enable the Reparameterization module (Sec.~\ref{sec:reparam}) to selectively route energy to the appropriate subspace based on video quality.

\subsection{Dual-Space Reparameterization}
\label{sec:reparam}
\begin{table*}[t]
    \centering
    \renewcommand{\arraystretch}{1.0}
    \setlength{\tabcolsep}{2.0pt}
    \caption{\textbf{Quantitative results on the Mixed-Quality VidSTG test set (Sec.~\ref{dataset})}.
We benchmark our NST against Full Fine-tune and advanced PEFT methods based on the same SOTA backbone~\cite{gu2024context}. 
\colorbox{green!10}{\textbf{Ours}} achieves the best trade-off between restoration and preservation, consistently surpassing other tuning strategies. }
    \resizebox{\textwidth}{!}{
        \begin{tabular}{lcccccccc}
            \specialrule{1.5pt}{0pt}{0pt}
            
            \rowcolor{gray!15}
             & \multicolumn{4}{c}{\textbf{Declarative Sentences}} & \multicolumn{4}{c}{\textbf{Interrogative Sentences}} \\
            
            \rowcolor{gray!15}
            \multirow{-2}{*}{\textbf{Method}} & m\_tIoU & m\_vIoU & vIoU\,@0.3 & vIoU\,@0.5 & m\_tIoU & m\_vIoU & vIoU\,@0.3 & vIoU\,@0.5 \\
            
            \hline
            \hline
            
            Baseline\pub{CVPR24}\cite{gu2024context} & 49.93 & 28.95 & 41.46 & 26.06 & 48.69 & 24.26 & 34.04 & 20.76 \\
            Full-Finetune & 49.42 & 29.13 & 41.74 & 26.31 & 48.38 & 24.38 & 34.42 & 21.03 \\
            LoRA \pub{ICLR22}~\cite{hu2022lora} & 49.61 & 29.47 & 42.03 & 26.74 & 48.54 & 25.03 & 34.97 & 21.51 \\
            AdaLoRA \pub{ICLR23}~\cite{zhang2023adalora} & 49.73 & \underline{29.93}& 42.31 & 27.08 & 48.67 & 25.29 & 35.26 & 21.84 \\
            DoRA \pub{ICML24}~\cite{liu2024dora}  & \underline{50.02} & 29.72  & 42.74 & \underline{27.51} & 48.81 & 25.54 & 35.63 & \underline{22.28} \\
            FlyLoRA \pub{NeurIPS25}~\cite{zou2025flylora} & 49.91 & 29.84 & \underline{42.97} & 27.36 & \underline{49.03} & \underline{25.67} & \underline{35.94} & 22.19 \\
            
            \hline
            
            \rowcolor{green!10}
            \textbf{Ours (NST)} & \textbf{50.81} \tiny{\textcolor{teal}{(+0.88)}} & \textbf{31.06} \tiny{\textcolor{teal}{(+2.11)}} & \textbf{44.11} \tiny{\textcolor{teal}{(+2.65)}} & \textbf{28.68} \tiny{\textcolor{teal}{(+2.62)}} & \textbf{49.47} \tiny{\textcolor{teal}{(+0.78)}} & \textbf{26.49} \tiny{\textcolor{teal}{(+2.23)}} & \textbf{36.90} \tiny{\textcolor{teal}{(+2.86)}} & \textbf{23.29} \tiny{\textcolor{teal}{(+2.53)}} \\
            
            \specialrule{1.5pt}{0pt}{0pt}
        \end{tabular}
    }
    
    \label{tab:vidstg_mixed_final}
    \vspace{-3mm}
\end{table*}

\begin{table*}[t]
    \centering
    \renewcommand{\arraystretch}{1.0}
    \setlength{\tabcolsep}{2.0pt}
    \caption{\textbf{Quantitative results of Weighted Adaptation Score based on VidSTG HQ and LQ test set (Sec.~\ref{dataset}).} Higher WAS scores ($\Delta_{LQ} - |\Delta_{HQ}|$) indicate better capability to restore low-quality inputs while preserving high-quality performance. \colorbox{green!10}{\textbf{Ours}} consistently achieves the highest WAS across all sentence types and metrics.}
    \resizebox{\textwidth}{!}{
    \begin{tabular}{lcccccccc}
        \specialrule{1.5pt}{0pt}{0pt}
        
        \rowcolor{gray!15}
         & \multicolumn{4}{c}{\textbf{Declarative Sentences}} & \multicolumn{4}{c}{\textbf{Interrogative Sentences}} \\
        
        \rowcolor{gray!15}
        \multirow{-2}{*}{\textbf{Method}} & $\Delta$m\_tIoU & $\Delta$m\_vIoU & $\Delta$vIoU\,@0.3 & $\Delta$vIoU\,@0.5 & $\Delta$m\_tIoU & $\Delta$m\_vIoU & $\Delta$vIoU\,@0.3 & $\Delta$vIoU\,@0.5 \\
        
        \hline
        \hline
        Full-Finetune & 0.12 & 0.56 & 1.24 & 0.83 & -0.05 & 0.42 & 1.11 & 0.97 \\
        LoRA\pub{ICLR22}\cite{hu2022lora} & 0.16 & 2.12 & 2.67 & 2.50 & -0.02 & 1.61 & 3.13 & 2.10 \\
        AdaLoRA\pub{ICLR23}\cite{zhang2023adalora} & 0.18 & 2.19 & 3.10 & 2.38 & -0.04 & 2.34 & 3.06 & 1.95 \\
        DoRA\pub{ICML24}\cite{liu2024dora} & 0.34 & 2.48 & 3.44 & 2.76 & 0.08 & 2.66 & 3.45 & 2.48 \\
        FlyLoRA\pub{NeurIPS25}\cite{zou2025flylora} & \underline{0.42} & \underline{2.63} & \underline{3.55} & \underline{3.07} & \underline{0.16} & \underline{2.75} & \underline{3.54} & \underline{2.65} \\
        
        \hline
        
        \rowcolor{green!10}
       \textbf{Ours (NST)} & \textbf{1.17} & \textbf{3.98} & \textbf{4.60} & \textbf{4.50} & \textbf{0.71} & \textbf{3.87} & \textbf{5.19} & \textbf{4.42} \\
        
        \specialrule{1.5pt}{0pt}{0pt}
    \end{tabular}
}
    
    
    \label{tab:vidstg_was}
    \vspace{-3mm}
\end{table*}

\begin{table}[t]
    \centering
    \setlength{\tabcolsep}{4.0pt} 
    \renewcommand{\arraystretch}{1.1} 
    \caption{\textbf{Ablation Study on Core Mechanisms.} 
    We report the Weighted Adaptation Score on HCSTVG-v1. 
    \textit{Optimization-based}: Geometric loss constraints. 
    \textit{Structural}: Our dual-space reparameterization.
    }
    \resizebox{\columnwidth}{!}{
        \begin{tabular}{cc|cccc}
            \specialrule{1.5pt}{0pt}{0pt}
            
            \rowcolor{gray!15}
            \textbf{QAU} & \textbf{Constraint} & \textbf{$\Delta$m\_tIoU} & \textbf{$\Delta$m\_vIoU} & \textbf{$\Delta$vIoU\,@0.3} & \textbf{$\Delta$vIoU\,@0.5} \\
            
            \hline
            \hline
            
            \xmark & Structural & -0.08 & +0.17 & +0.23 & +0.09 \\
            
            \cmark & $-$ & +0.15 & +0.36 & +0.42 & +0.31 \\
            
            \cmark & Optimization-based & +0.85 & +2.66 & +3.15 & +2.88 \\
            
            \rowcolor{green!10} 
            \cmark & \textbf{Structural} & \textbf{+1.25} & \textbf{+4.18} & \textbf{+4.71} & \textbf{+4.53} \\
            
            \specialrule{1.5pt}{0pt}{0pt}
        \end{tabular}
    }

    \label{tab:ablation_was}
    \vspace{-3mm}
\end{table}

\begin{table}[t]
    \centering
    \setlength{\tabcolsep}{4.0pt} 
    \renewcommand{\arraystretch}{1.1}   
    \caption{\textbf{Analysis on Layer Selection Strategy.} 
    We report the Weighted Adaptation Score on HCSTVG-v1.
    \textit{Low/High-Nullity}: Applying our NST tuning framework to layers with minimal/maximal null-space capacity(full/low rank).
    }
    \resizebox{\columnwidth}{!}{
        \begin{tabular}{l|cccc}
            \specialrule{1.5pt}{0pt}{0pt}
            
            \rowcolor{gray!15}
            \textbf{Strategy} & \textbf{$\Delta$m\_tIoU} & \textbf{$\Delta$m\_vIoU} & \textbf{$\Delta$vIoU\,@0.3} & \textbf{$\Delta$vIoU\,@0.5} \\
            
            \hline
            \hline
            
            Low-Nullity & +0.12 & +0.35 & +0.41 & +0.28 \\
            
            Random Selection & +0.65 & +1.95 & +2.17 & +2.13 \\
            
            \rowcolor{green!10}
            \textbf{High-Nullity} & \textbf{+1.25} & \textbf{+4.18} & \textbf{+4.71} & \textbf{+4.53} \\
            
            \specialrule{1.5pt}{0pt}{0pt}
        \end{tabular}
    }

    \label{tab:ablation_layers_flat}
    \vspace{-3mm}
\end{table}
With the projection coefficients $\vec{\alpha}$ and $\vec{\beta}$ predicted by the QAU, we propose a geometric mechanism to generate the final residual $\delta$. As depicted in Figure \ref{fig:framework}(c), this module leverages the pre-trained weight to handle HQ inputs via null-space and LQ inputs via row-space, which effectively decouples feature restoration from knowledge preservation.

Specifically, we perform Singular Value Decomposition(SVD) on the frozen weights $\mathbf{W} = \mathbf{U}\mathbf{\Sigma}\mathbf{V}^\top$ to 
separate the feature space into two orthogonal bases $\mathbf{V} = [\mathbf{V}_{row} \mid \mathbf{V}_{null}]$. Here, $\mathbf{V}_{row}$ spans the Row-Space, while $\mathbf{V}_{null}$ spans the Null-Space. This decomposition is pre-computed and cached, incurring negligible overhead. To dynamically select the appropriate subspace, we introduce a lightweight router that estimates the noise level from the semantic gap $\mathbf{H}$, predicting a gate $s = \sigma(\text{MLP}_{gate}(\text{AvgPool}(\mathbf{H}))) \in [0,1]$.
Based on this gate, we construct the residual $\delta \in \mathbb{R}^{N \times D}$ via orthogonal projection:
\begin{equation}
    \delta = \underbrace{s \cdot (\vec{\beta} \mathbf{V}_{row}^\top)}_{\text{Restoration}} + \underbrace{(1-s) \cdot (\vec{\alpha} \mathbf{V}_{null}^\top)}_{\text{Preservation}}.
    \label{eq:hard_reparam}
\end{equation}

For HQ inputs, the router suppresses the row-space term. The residual becomes dominated by the null-space term $\delta \approx \vec{\alpha}\mathbf{V}_{null}^\top$. Since $\mathbf{W}\mathbf{V}_{null} = \mathbf{0}$, this residual is effectively nullified in the layer output, preserving pre-trained features. For LQ inputs, the router activates the row-space term. The residual becomes $\delta \approx \vec{\beta}\mathbf{V}_{row}^\top$. Since $\mathbf{W}\mathbf{V}_{row} \neq \mathbf{0}$, this component effectively passes through the weights to rectify the degraded features. The features are then updated as $\hat{\mathbf{F}} = \mathbf{F} + \delta$. During the forward propagation, the frozen backbone $\mathbf{W}$ mathematically eliminates the null-space component ($\mathbf{W}\delta = \mathbf{0}$) to preserve original outputs, while allowing the row-space component to pass through for restoration. 
\vspace{-1mm}
\subsection{Optimization}
\begin{figure}[t!]
    \centering 
    \includegraphics[width=\linewidth]{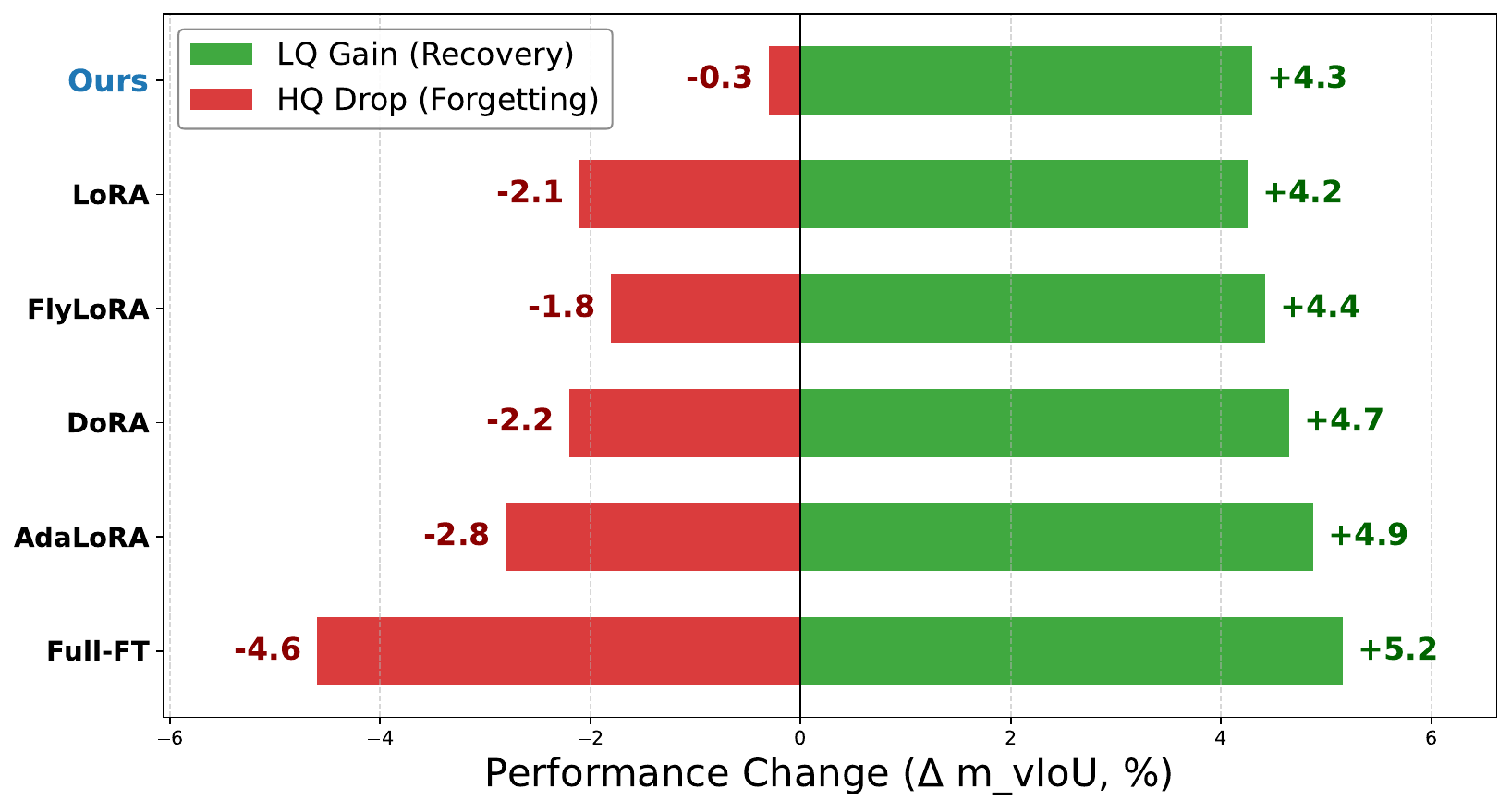}
    \caption{\textbf{Restoration vs. Preservation Trade-off on VidSTG.} 
    We visualize the performance change on $\Delta$m\_vIoU relative to the Zero-shot Baseline. 
    \textcolor{gaincolor}{\textbf{Green bars}} indicate gains on LQ data , while \textcolor{dropcolor}{\textbf{Red bars}} indicate drops on HQ data.}
    \label{fig:tradeoff}
    \vspace{-5mm}
\end{figure}

The training objective combines the standard grounding loss $\mathcal{L}_{STVG}$ with a router supervision term and a subspace regularization term:
\begin{equation}
    \mathcal{L}_{total} = \mathcal{L}_{STVG} + \lambda_{gate}\mathcal{L}_{gate}  + \lambda_{reg}\mathcal{L}_{reg}.
\end{equation}

Specifically, $\mathcal{L}_{gate}=\| s - y_{q} \|^2$ utilizes quality labels $y_{q}$($0$ for HQ, $1$ for LQ) to supervise the router. $\mathcal{L}_{reg}$ regulates the projection coefficients based on the input quality label $y_q$ :
\begin{equation}
    \mathcal{L}_{reg} = (1-y_q) \cdot \|\vec{\beta}\|_2^2 + y_q \cdot \|\vec{\alpha}\|_2^2.
\end{equation}

For HQ inputs, we minimize $\|\vec{\beta}\|_2^2$ to suppress the row-space component, mitigating the risk of feature distortion. For LQ inputs, minimizing $\|\vec{\alpha}\|_2^2$ is numerically essential: since the router s applies soft gating, this constraint prevents magnitude explosion that would cause noise leakage. Furthermore, it enforces energy concentration, compelling the model to channel all valid restoration signals into the row-space via $\vec{\beta}$ while restricting $\vec{\alpha}$ to absorb stochastic noise.
\begin{figure*}[t!]
    \centering 
    \includegraphics[width=\linewidth]{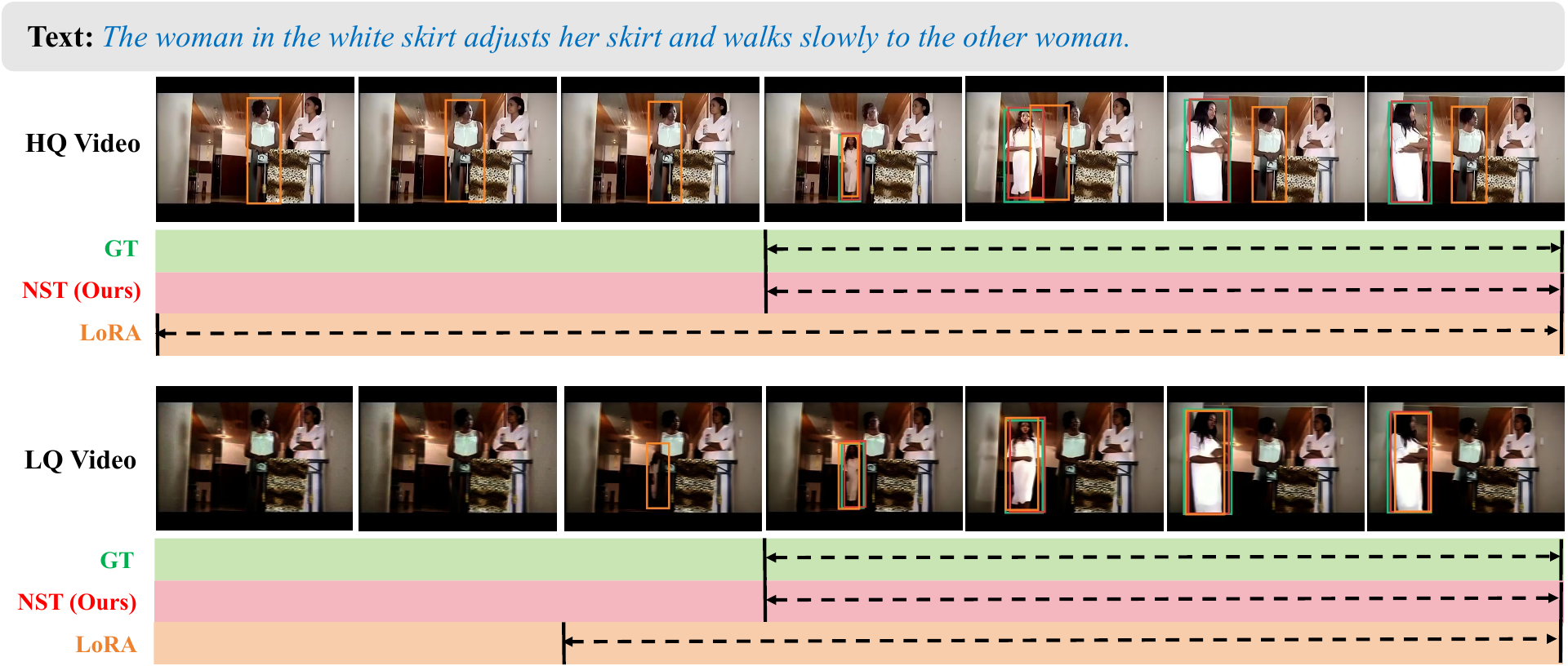}
    \vspace{-5mm}
    \caption{\textbf{Qualitative comparison on mixed-quality inputs.} LoRA suffers forgetting on HQ inputs (Top) and misses fine-grained cues on LQ inputs (Bottom). In contrast, NST maintains precision on both by restoring LQ inputs while preserving HQ ones.
    }
    \label{result}
    \vspace{-5mm}
\end{figure*}

\section{Results and Analysis}
\noindent\textbf{Implementation details.}
Based on CG-STVG~\cite{gu2024context}, we use ResNet-101~\cite{he2016deep} and VidSwin-T~\cite{liu2022video} for visual feature extraction and RoBERTa-base~\cite{liu2019roberta} for text encoding. The visual and textual backbones are initialized from MDETR~\cite{kamath2021mdetr} and kept frozen during training. We use the AdamW optimizer~\cite{loshchilov2017decoupled} with a weight decay of $1e-4$ and a learning rate of $3e-4$ for NST parameters. Loss weights are set to $\lambda_{gate}=5, \lambda_{reg}=1$. All experiments are conducted on NVIDIA RTX 5880 Ada Generation GPUs with a batch size of 32.

\subsection{Datsets and Metrics}
\label{dataset}
\noindent\textbf{Datasets.} 
We use two common used benchmarks: HCSTVG-v1/v2~\cite{tang2021human} and VidSTG~\cite{where}. Following~\cite{embrace, lin2023collaborative, gu2024context}, we report results on the validation sets for HCSTVG (v1: 1,160, v2: 2,000 samples) and the test set for VidSTG (10,303 samples). To better benchmark robustness against real-world quality shifts, we design a physics-based degradation pipeline $\Phi(\cdot)$. We apply this pipeline to the datasets to generate their Low-Quality counterparts. By integrating these LQ samples with the original HQ data at a 1:1 ratio, we construct a \textbf{Mixed-Quality Benchmark} to simultaneously evaluate the model's restoration capability on LQ inputs and preservation ability on HQ ones. Specifically, $\Phi(\cdot)$ comprises:
(1) Motion Blur via randomly generated directional kernels simulating rapid camera shake (size $\in [3, 21]$, angle $\in [0, 360^\circ]$); 
(2) Defocus Blur via Gaussian kernels with varying standard deviations to simulate optical focus loss; 
(3) Low-Light simulation via linear brightness reduction and nonlinear Gamma correction for underexposed scenes; 
and (4) Resolution degradation by downsampling to discard fine-grained details.

\noindent\textbf{Metrics.} 
Following prior works~\cite{yang2022tubedetr, gu2024context}, we employ m\_tIoU, m\_vIoU, and vIoU\,@R as evaluation metrics. m\_tIoU measures the average temporal Intersection-over-Union (IoU) between predicted and ground-truth segments. 
m\_vIoU evaluates the spatio-temporal accuracy by averaging spatial IoUs over the union of temporal frames. 
vIoU\,@R denotes the percentage of test samples where vIoU exceeds a threshold $R \in \{0.3, 0.5\}$. Furthermore, to better evaluate the trade-off between restoration and preservation, we propose the \textbf{Weighted Adaptation Score} metric:
\begin{equation}
    \text{WAS} = \Delta_{\text{LQ}} - |\Delta_{\text{HQ}}|,
    \label{was}
\end{equation}
where $\Delta_{\text{LQ}}$ is the performance gain on LQ data and $\Delta_{\text{HQ}}$ is the performance drop on HQ data relative to the zero-shot baseline. A higher WAS indicates effective restoration with minimal catastrophic forgetting.
\subsection{State-of-the-art Comparison}
\label{sec:comparison}
To validate the effectiveness and robustness of NST, we benchmark it against Full Fine-tune and state-of-the-art PEFT methods based on~\cite{gu2024context}. Evaluation is conducted across three versions of the VidSTG test set: High-Quality, Low-Quality, and Mixed-Quality. The rank $r$ is set to 64 for all PEFT methods. As detailed in Table~\ref{tab:vidstg_mixed_final}, NST consistently outperforms all methods on the Mixed-Quality set. Specifically, it surpasses Full Fine-tune by 1.39\% in m\_tIoU and 2.37\% in vIoU\,@0.5 for declarative sentences, demonstrating robust adaptability to complex distribution shifts in mixed-quality scenarios. To quantify the balance between restoration and preservation, we calculate the Weighted Adaptation Score based on HQ and LQ results. As shown in Table~\ref{tab:vidstg_was}. NST achieves the highest WAS across all sentence types and metrics (e.g., 4.50\% on vIoU\,@0.5), outperforming other tuning strategies. To better visualize this superiority, we plot the performance trade-off in Fig.~\ref{fig:tradeoff}. As shown, Full Fine-tune suffers severe catastrophic forgetting with a 4.6\% drop on HQ. Similarly, AdaLoRA yields a 4.9\% gain on LQ inputs but degrades HQ results with a 2.8\% drop. In contrast, our method achieves a balance by matching the restoration capability of leading PEFT methods with a 4.3\% on LQ while maintaining HQ performance with a  0.3\% drop. This confirms that our method successfully refines degraded inputs while preserving pre-trained knowledge.

\subsection{Ablation Study}
\label{sec:ablation}

\vspace{1mm}
\noindent\textbf{Impact of Core Mechanisms.} 
To validate our design, we conduct ablation studies on HCSTVG-v1 and report the WAS across metrics. As shown in Table~\ref{tab:ablation_was}, replacing the QAU with a simple MLP with only visual features yields the poorest performance, which emphasizes the critical importance of the QAU. However, even with semantic guidance from QAU, naive injection without constraints yields a negligible WAS, as blindly injecting residuals severely disrupts the performance on HQ data. 
Although the optimization-based strategy improves the score via loss regularization, it still fails to enforce strict orthogonality solely through optimization. In contrast, our structural reparameterization achieves the highest WAS. 
This comparison demonstrates that our design effectively optimizes the trade-off between restoration and preservation.

\vspace{1mm}
\noindent\textbf{Impact of Matrix Null-Space Capacity.}
To validate the necessity of geometric redundancy, we evaluate layer selection strategies based on the intrinsic rank of pre-trained weights. We categorize layers into High-Nullity layers with lowest 10\% in rank ratio and Low-Nullity layers with highest 10\% in rank ratio. We report the WAS on the HCSTVG-v1 validation set. As shown in Table~\ref{tab:ablation_layers_flat}, inserting modules into Low-Nullity (full rank) layers yields a worst WAS of +0.09\% on vIoU\,@0.5. The limited null-space in these layers creates a bottleneck, as it lacks sufficient capacity to contain potential noise. Consequently, this noise interferes with the processing of HQ data. In contrast, the High-Nullity (low rank) strategy achieves a superior WAS of +4.53\%. This confirms that using layers with larger null-space capacity is essential, as they provide sufficient orthogonal degrees of freedom to isolate noise for HQ data while enabling restoration for LQ inputs.

\subsection{Qualitative Analysis}
\label{sec:qualitative}
We present qualitative results in Fig.~\ref{result}. On HQ inputs, LoRA exhibits imprecise temporal boundaries for the action “walk”, suggesting interference with the pre-trained knowledge. In contrast, NST preserves the original knowledge, yielding precise boundaries. On LQ inputs, while LoRA localizes the target, it produces misaligned bounding boxes for the fine-grained query “adjusts her skirt”, indicating difficulty in aligning degraded features with subtle semantics. Conversely, NST successfully rectifies these features, resulting in precise localization. This comparison shows that our framework effectively balances feature restoration with knowledge preservation.

\section{Conclusions}
In this work, we introduced Null-Space Tuning to resolve the fundamental dilemma between adapting to mixed-quality inputs and preserving pre-trained knowledge in real-world STVG. Our key innovation lies in geometrically decoupling adaptation updates: routing restoration signals to the row-space for degraded inputs while directing potential noise for HQ data within the null-space to make it invisible to the backbone. While demonstrating superior robustness, we note that NST is primarily tailored for linear projection layers, relying on the existence of sufficient geometric redundancy in pre-trained weights to harbor these adaptation residuals.

\section*{Acknowledgment}
This work was supported partially by the NSFC(62476296), Guangdong Natural Science Funds Project (2023B1515040025, 2022B1111010002, 2024A1111120017), Guangdong NSF for DistinguishedYoung Scholar (2022B1515020009).

\bibliographystyle{IEEEbib}
\bibliography{icme2026references}

@inproceedings{where,
  title={Where does it exist: Spatio-temporal video grounding for multi-form sentences},
  author={Zhang, Zhu and others},
  booktitle={CVPR},
  pages={10668--10677},
  year={2020}
}

@article{embrace,
  title={Embracing consistency: A one-stage approach for spatio-temporal video grounding},
  author={Jin, Yang and Yuan, Zehuan and others},
  journal={NeurIPS},
  volume={35},
  pages={29192--29204},
  year={2022}
}

@inproceedings{lin2023collaborative,
  title={Collaborative static and dynamic vision-language streams for spatio-temporal video grounding},
  author={Lin, Zihang and Tan, Chaolei and others},
  booktitle={CVPR},
  pages={23100--23109},
  year={2023}
}

@inproceedings{yang2022tubedetr,
  title={Tubedetr: Spatio-temporal video grounding with transformers},
  author={Yang, Antoine and Miech, Antoine and others},
  booktitle={CVPR},
  pages={16442--16453},
  year={2022}
}

@inproceedings{carion2020end,
  title={End-to-end object detection with transformers},
  author={Carion, Nicolas and Massa, Francisco and others},
  booktitle={ECCV},
  pages={213--229},
  year={2020},
  organization={Springer}
}

@inproceedings{fang2024alphaedit,
  author       = {Junfeng Fang and
                  Houcheng Jiang and
                  others},
  title        = {AlphaEdit: Null-Space Constrained Knowledge Editing for Language Models},
  booktitle    = {ICLR},
  publisher    = {OpenReview.net},
  year         = {2025},
  url          = {https://openreview.net/forum?id=HvSytvg3Jh},
  bibsource    = {dblp computer science bibliography, https://dblp.org}
}

@article{zhang2020object,
  title={Object-aware multi-branch relation networks for spatio-temporal video grounding},
  author={Zhang, Zhu and Zhao, Zhou and others},
  journal={arXiv preprint arXiv:2008.06941},
  year={2020}
}

@inproceedings{kamath2021mdetr,
  title={Mdetr-modulated detection for end-to-end multi-modal understanding},
  author={Kamath, Aishwarya and others},
  booktitle={ICCV},
  pages={1780--1790},
  year={2021}
}

@article{loshchilov2017decoupled,
  title={Decoupled weight decay regularization},
  author={Loshchilov, Ilya and Hutter, Frank},
  journal={arXiv preprint arXiv:1711.05101},
  year={2017}
}

@inproceedings{wang2021training,
  title={Training networks in null space of feature covariance for continual learning},
  author={Wang, Shipeng and Li, Xiaorong and others},
  booktitle={CVPR},
  pages={184--193},
  year={2021}
}

@article{cheng2024mamba,
  title={Mamba-CL: Optimizing Selective State Space Model in Null Space for Continual Learning},
  author={Cheng, De and Lu, Yue and others},
  journal={arXiv preprint arXiv:2411.15469},
  year={2024}
}

@inproceedings{jia2022visual,
  title={Visual prompt tuning},
  author={Jia, Menglin and Tang, Luming and others},
  booktitle={ECCV},
  pages={709--727},
  year={2022},
  organization={Springer}
}

@inproceedings{zhang2023adalora,
  author       = {Qingru Zhang and
                  Minshuo Chen and
                  others},
  title        = {Adaptive Budget Allocation for Parameter-Efficient Fine-Tuning},
  booktitle    = {ICLR},
  publisher    = {OpenReview.net},
  year         = {2023},
  url          = {https://openreview.net/forum?id=lq62uWRJjiY},
  bibsource    = {dblp computer science bibliography, https://dblp.org}
}

@inproceedings{houlsby2019parameter,
  title={Parameter-efficient transfer learning for NLP},
  author={Houlsby, Neil and Giurgiu, Andrei and others},
  booktitle={ICML},
  pages={2790--2799},
  year={2019},
  organization={PMLR}
}

@inproceedings{su2021stvgbert,
  title={Stvgbert: A visual-linguistic transformer based framework for spatio-temporal video grounding},
  author={Su, Rui and others},
  booktitle={ICCV},
  pages={1533--1542},
  year={2021}
}

@inproceedings{gu2024context,
  title={Context-guided spatio-temporal video grounding},
  author={Gu, Xin and Fan, Heng and others},
  booktitle={CVPR},
  pages={18330--18339},
  year={2024}
}

@inproceedings{gu2025knowing,
  author       = {Xin Gu and
                  Yaojie Shen and
                  others},
  title        = {Knowing Your Target: Target-Aware Transformer Makes Better Spatio-Temporal
                  Video Grounding},
  booktitle    = {ICLR},
  publisher    = {OpenReview.net},
  year         = {2025},
  url          = {https://openreview.net/forum?id=WOzffPgVjF},
  bibsource    = {dblp computer science bibliography, https://dblp.org}
}

@article{hu2022lora,
  title={Lora: Low-rank adaptation of large language models},
  author={Hu, Edward J and Shen, Yelong and others},
  journal={ICLR},
  volume={1},
  number={2},
  pages={3},
  year={2022}
}

@article{tang2021human,
  title={Human-centric spatio-temporal video grounding with visual transformers},
  author={Tang, Zongheng and Liao, Yue and others},
  journal={IEEE TCSVT},
  volume={32},
  number={12},
  pages={8238--8249},
  year={2021},
  publisher={IEEE}
}

@inproceedings{liu2024dora,
  title={Dora: Weight-decomposed low-rank adaptation},
  author={Liu, Shih-Yang and Wang, Chien-Yi and others},
  booktitle={ICML},
  year={2024}
}

@inproceedings{he2016deep,
  title={Deep residual learning for image recognition},
  author={He, Kaiming and Zhang, Xiangyu and others},
  booktitle={CVPR},
  pages={770--778},
  year={2016}
}

@inproceedings{liu2022video,
  title={Video swin transformer},
  author={Liu, Ze and others},
  booktitle={CVPR},
  pages={3202--3211},
  year={2022}
}

@article{liu2019roberta,
  title={Roberta: A robustly optimized bert pretraining approach},
  author={Liu, Yinhan and Ott, Myle and others},
  journal={arXiv preprint arXiv:1907.11692},
  year={2019}
}

@article{zou2025flylora,
  title={FlyLoRA: Boosting Task Decoupling and Parameter Efficiency via Implicit Rank-Wise Mixture-of-Experts},
  author={Zou, Heming and Zang, Yunliang and others},
  journal={arXiv preprint arXiv:2510.08396},
  year={2025}
}
\end{document}